\documentclass[10pt,twocolumn,letterpaper]{article}

\usepackage[pagenumbers]{cvpr} % To force page numbers, e.g. for an arXiv version

\definecolor{cvprblue}{rgb}{0.21,0.49,0.74}
\usepackage[pagebackref,breaklinks,colorlinks,allcolors=cvprblue]{hyperref}
\usepackage{booktabs}
\usepackage{multirow}
\usepackage{makecell}
\usepackage{array}
\usepackage{xcolor} % 颜色支持包
\def\paperID{19878} % *** Enter the Paper ID here
\def\confName{CVPR}
\def\confYear{2026}

\title{MFI-ResNet: Efficient ResNet Architecture Optimization via MeanFlow Compression and Selective Incubation}

\author{
Nuolin Sun\\
Linyuan Wang\\
Haonan Wei\\
Lei Li$^{*}$\\
Bin Yan\\
Information Engineering University\\
{\tt\small sunnuolinsnl@163.com}
\thanks{$^{*}$Corresponding author.}
}

\begin{document}
\maketitle
\begin{abstract}
ResNet has achieved tremendous success in computer vision through its residual connection mechanism. ResNet can be viewed as a discretized form of ordinary differential equations (ODEs). From this perspective, the multiple residual blocks within a single ResNet stage essentially perform multi-step discrete iterations of the feature transformation for that stage. The recently proposed flow matching model, MeanFlow, enables one-step generative modeling by learning the mean velocity field to transform distributions. Inspired by this, we propose MeanFlow-Incubated ResNet (MFI-ResNet), which employs a compression-expansion strategy to jointly improve parameter efficiency and discriminative performance. In the compression phase, we simplify the multi-layer structure within each ResNet stage to one or two MeanFlow modules to construct a lightweight meta model. In the expansion phase, we apply a selective incubation strategy to the first three stages, expanding them to match the residual block configuration of the baseline ResNet model, while keeping the last stage in MeanFlow form, and fine-tune the incubated model. Experimental results show that on CIFAR-10 and CIFAR-100 datasets, MFI-ResNet achieves remarkable parameter efficiency, reducing parameters by 46.28\% and 45.59\% compared to ResNet-50, while still improving accuracy by 0.23\% and 0.17\%, respectively. This demonstrates that generative flow-fields can effectively characterize the feature transformation process in ResNet, providing a new perspective for understanding the relationship between generative modeling and discriminative learning.
\end{abstract}    
\section{Introduction}
\label{sec:intro}

Since its proposal in 2016, ResNet has effectively addressed the deep network degradation problem through its residual learning mechanism and skip connection design, becoming a foundational architecture in the field of computer vision\cite{He1}. ResNet has demonstrated outstanding performance in fundamental vision tasks such as image classification\cite{He1,szegedy6,he7,hu8}, object detection\cite{lin2,tan3}, and semantic segmentation\cite{Zhao4,Chen5}, as well as in cutting-edge areas including adversarial robustness\cite{madry2018towards,zhang2019theoretically,wang2019improving} and multimodal learning\cite{radford2021learning,lu2019vilbert}. Ongoing research into ResNet's architecture, connection patterns, and training strategies continues to drive improvements in deep model performance.

Recent theoretical research in deep learning has revealed a profound connection between ResNet and ordinary differential equations (ODEs)\cite{haber2017stable,lu2018beyond,chen2018neural}. From this perspective, a residual connection is essentially an Euler discretization of an ODE. The network depth corresponds to the time dimension, and the function modeled by each residual block characterizes the instantaneous velocity field of the feature evolution. Consequently, the core computation of each residual block in a ResNet can be seen as solving for the instantaneous velocity corresponding to the input features at a discrete time step. The entire network then transforms source features into target features through the sequential accumulation of these instantaneous velocities in a multi-step iterative manner, a process whose efficiency is limited by the computational overhead of these multiple steps.

Concurrently, generative models have achieved breakthrough progress in computer vision\cite{ho2020denoising,goodfellow2014generative,rombach2022high,dhariwal2021diffusion,song2019generative}. In particular, probability flow ODE–based generative models describe feature evolution by solving an ODE system that defines a continuous path from a noise distribution to a data distribution\cite{lipman2022flow,tong2023improving}. The core of these models lies in learning an instantaneous velocity field and integrating it over time to drive samples along this path. From a mathematical standpoint, both flow models and ResNets can be viewed as learning velocity fields in feature space. In ResNet, each residual block estimates an instantaneous velocity whose accumulation through residual connections yields feature transformations for discrimination, whereas in flow models, the estimated velocity field transports probability mass for generation.

\begin{figure*}[t]
  \centering
  \includegraphics[width=1.0\linewidth]{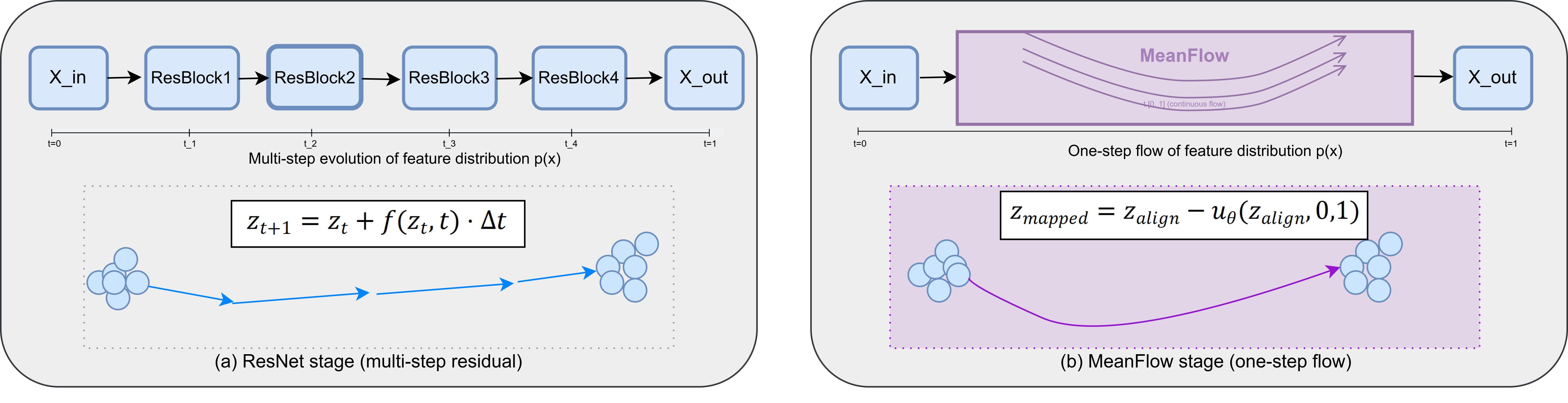}
   \caption{Comparison of feature transformation in a ResNet stage and a MeanFlow stage. (a) A ResNet stage implements a multi-step residual evolution of the feature distribution, where several residual blocks realize the discrete update. (b) A MeanFlow stage models a one-step continuous flow of the feature distribution, directly mapping via an average velocity field.}
   \label{fig:meanflow}
\end{figure*}

Recently, the MeanFlow method proposed by He et al. introduced the concept of an average velocity field, providing a principled and effective framework for single-step generative modeling\cite{geng2025mean}. The central idea of this method is to shift from solving for the instantaneous velocity field through multi-step iteration to directly modeling the average velocity field across a time interval. This shift enables the model to complete the mapping from noise to data in a single computational step, avoiding the multi-step numerical integration required by traditional methods.

This naturally leads to a question: since residual blocks in ResNet learn a discretized instantaneous velocity field, while MeanFlow achieves a single-step mapping via an average velocity field, can the multi-step computation in ResNet be replaced by MeanFlow's average-velocity formulation? As illustrated in Fig.~\ref{fig:meanflow}, from an ODE perspective, a continuous feature transformation flow requires a fixed-dimensional domain. The stage-wise design of ResNet provides exactly such a structure, as the feature map dimensions remain constant within each stage. This allows us to reinterpret the accumulation of multiple discrete instantaneous velocities within a stage as the integration of an underlying continuous velocity field. Following the core idea of MeanFlow, we further directly model the average velocity over the whole transformation interval. We therefore retain ResNet's original four-stage hierarchy but replace, within each stage, the multi-step accumulation with a single-step or few-step explicit transformation modeled by MeanFlow. In this way, we substantially reduce the model depth while preserving the benefits of the stage-wise organization. However, this compression inevitably reduces model capacity, making it necessary to design an effective expansion mechanism to maintain discriminative power.

Deep Incubation provides a feasible implementation route for this purpose~\cite{ni2023deep}. It first pre-trains a lightweight meta model as the training backbone, and then embeds each target module into the meta model and trains it independently while keeping the other modules fixed, thereby improving performance while reducing training cost. Inspired by this idea, we propose a compression-expansion strategy. In the compression phase, we employ MeanFlow to construct a lightweight meta model and thus compress the network. In the expansion phase, we observe a pronounced imbalance in the parameter distribution across ResNet stages, where later stages contain far more parameters than earlier ones. This observation motivates us to borrow the spirit of Deep Incubation and adopt a selective incubation strategy that treats different stages differently, so as to recover key discriminative features while maximally preserving the parameter savings brought by compression.

Based on the preceding analysis, we propose MeanFlow-Incubated ResNet (MFI-ResNet), a ResNet optimization method that compresses stage-wise residual blocks into MeanFlow modules and then selectively incubates them with pre-trained ResNet stages. This method employs a two-stage "compression-expansion" strategy, aiming to synergistically enhance both parameter efficiency and discriminative performance. We first leverage MeanFlow to construct and compress a lightweight meta model, which is then expanded using a selective incubation strategy. Experiments on the CIFAR-10 and CIFAR-100 datasets demonstrate that compared to the baseline ResNet-50, MFI-ResNet achieves a 0.23\% and 0.17\% improvement in classification accuracy, respectively, while reducing the parameter count by 46.28\% and 45.59\%. These results indicate that generative flow-fields can effectively characterize the feature transformation process within ResNet, offering a new perspective on the intrinsic connection between discriminative learning and generative modeling.
\section{Related Works}
\label{sec:Related Works}

\subsection{The ODE View of ResNet and Flow Models}

The connection between ResNet and ODEs is well-established\cite{chen2018neural,dupont2019augmented,haber2017stable}. A ResNet's residual connection can be viewed as the Euler discretization for solving an ODE, where network layers correspond to discrete time steps and the residual function represents the instantaneous velocity field\cite{lu2020finitelayerneuralnetworks,chen2018neural}. From this perspective, each ResNet layer computes the instantaneous velocity at the current moment, achieving feature transformation through layer-by-layer accumulation. Similarly, generative models based on flow matching\cite{lipman2022flow,tong2023improving} also learn an instantaneous velocity field and solve an ODE to transform a noise distribution into a data distribution. MeanFlow proposes single-step generation by modeling an average velocity field\cite{geng2025mean}. Traditional flow matching requires estimating the instantaneous velocity at each time step and performing multi-step integration for sampling, which is computationally expensive\cite{song2020score,karras2022elucidating}. MeanFlow introduces the concept of an average velocity field, converting the multi-step integration of instantaneous velocities into a single-step, explicit calculation of the average velocity. This enables an equivalent distributional transformation via a single-step mapping. By deriving the differential relationship between average and instantaneous velocities, this method establishes an integration-free training objective, converting a multi-step numerical solving problem into an explicit algebraic one, thereby significantly improving both inference efficiency and training stability.

Based on this connection, our work introduces MeanFlow into the ResNet architecture for optimization. While the multi-layer residual blocks in each ResNet stage perform feature transformation through multi-step discretization, MeanFlow can achieve an equivalent transformation with fewer layers by using an average velocity field. We replace the stack of residual blocks in each stage with one or two MeanFlow modules, compressing the multi-layer structure into a few-layer flow-field module and achieving a substantial reduction in model depth.

%-------------------------------------------------------------------------
\subsection{Deep Incubation}

Training large-scale deep networks poses challenges of high computational cost and long training times\cite{tan2019efficientnet,han2015deep}. Deep Incubation proposes a modular training strategy to address this issue\cite{ni2023deep}. The method begins by pre-training a lightweight meta model, which has the same number of modules as the target model but with only a few layers per module. Then, during the incubation phase, each target module is sequentially embedded into the meta model and trained end-to-end while the other modules remain frozen. This allows the target module to learn task-oriented discriminative representations under the guidance of global semantics. Finally, the trained modules are assembled and fine-tuned to form the final model. Compared to traditional end-to-end training, this strategy not only significantly reduces training time but also improves the model's generalization performance and classification accuracy.

Our work draws inspiration from the Deep Incubation philosophy, using a MeanFlow-compressed lightweight model as the meta model and selectively incubating discriminative modules from a pre-trained ResNet. Unlike the original Deep Incubation, we employ a selective incubation strategy based on the parameter distribution characteristics of ResNet. We apply differential treatment to different stages to achieve an optimal balance between parameter efficiency and discriminative performance. The specific design details are elaborated in Section~\ref{sec:method32}.
\section{Method}

From an ODE perspective, the multi-layer residual blocks within each stage of a ResNet achieve feature transformation by accumulating multiple discrete instantaneous velocity fields. The theory of MeanFlow, however, offers a more efficient alternative. By directly modeling the average velocity field, an equivalent transformation can be accomplished through a single-step or few-step explicit mapping. This principle forms the theoretical basis for our compression of ResNet. Based on this, we propose MFI-ResNet, which employs a two-stage "compression-expansion" strategy. In the compression stage, we leverage MeanFlow to replace the multi-layer residual accumulation in ResNet with 1-2 layer flow-field modules, thereby constructing a lightweight meta model. However, this drastic reduction in depth leads to a significant decrease in model capacity, resulting in the loss of rich, hierarchical discriminative information from the original model. Therefore, in the expansion stage, we adopt a selective incubation strategy to restore performance by incorporating discriminative information from a pre-trained ResNet. In the following sections, we will elaborate on the module-level MeanFlow mapping for the compression stage (Section~\ref{sec:method31}) and the selective incubation for the expansion stage (Section~\ref{sec:method32}).

\begin{figure}[t]
  \centering
  \includegraphics[width=0.9\linewidth]{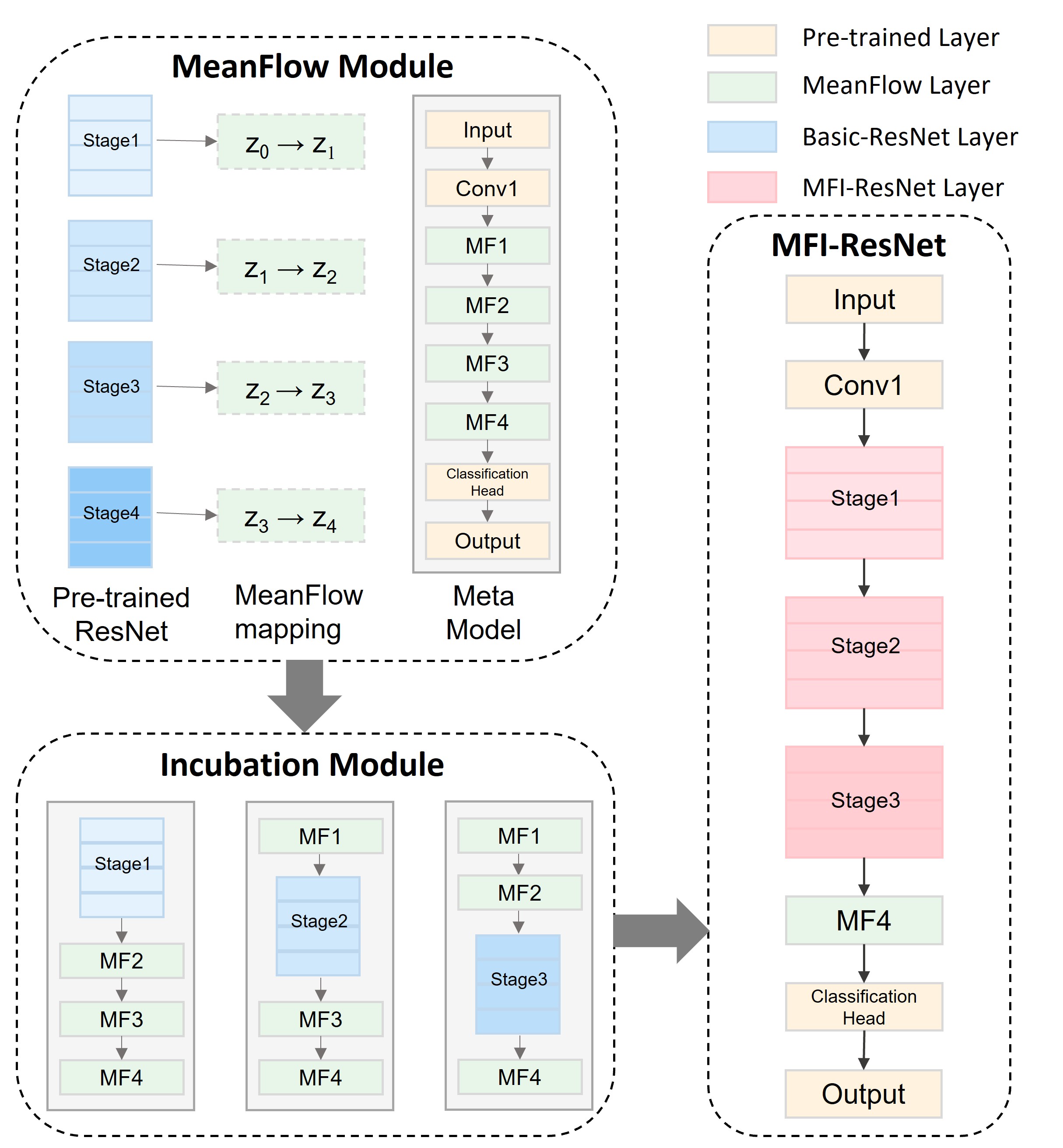}
   \caption{The MFI-ResNet architecture. \textbf{Top-left:} We train four independent MeanFlow modules to learn explicit feature mappings from pre-trained ResNet stages, constructing a lightweight meta model. \textbf{Bottom-left:} The selective incubation strategy progressively replaces MeanFlow modules with corresponding ResNet residual blocks for stages 1-3. \textbf{Right:} The final MFI-ResNet hybrid architecture combines incubated ResNet stages (1-3) with the retained MeanFlow module (stage 4), achieving parameter efficiency while maintaining discriminative performance.}
   \label{fig:framework}
\end{figure}

\subsection{Stage-level MeanFlow Mapping}
\label{sec:method31}
The multi-layer residual blocks within each stage of a ResNet can be viewed as a discrete form of an ODE. Specifically, a residual connection is essentially an Euler discretization of an ODE, where network depth corresponds to the time dimension, and the residual function corresponds to the instantaneous velocity field. From an ODE perspective, ResNet achieves the transformation from source features to target features through multi-step discrete iterations: \begin{equation} z_{t+1} = z_t + f(z_t, t) \cdot \Delta t \end{equation} where $f(\cdot)$ is the instantaneous velocity field modeled by the residual function, and $\Delta t$ is the time step. This design approximates a continuous feature evolution process by accumulating instantaneous velocities layer by layer, but it requires stacking a large number of layers to achieve a precise feature transformation.

MeanFlow, as a generative model based on flow matching, achieves an equivalent distributional transformation in a single forward pass by learning an average velocity field. Inspired by this, we utilize MeanFlow to learn the mapping from source to target features through continuous-time flow matching. Given a source feature distribution $z_{\text{align}}$ and a target feature distribution $z_{\text{target}}$, MeanFlow learns a continuous-time transformation driven by an ODE:
\begin{equation}
\frac{dz(t)}{dt} = u_\theta (z(t), t), \quad z(0) = z_{\text{align}}, \quad z(1) = z_{\text{target}}
\label{eq:emc}
\end{equation}
where $u_\theta(z(t), t)$ is a velocity field parameterized by a neural network.

MeanFlow learns the velocity field $u_\theta$ such that the features obtained after integrating the ODE approximate the target features $z_{\text{target}}$. This requires the source and target features to be fixed. We employ a pre-trained ResNet model as a fixed feature extractor to ensure the stability of the flow matching training.

For the $l$-th stage of a pre-trained ResNet model, let its input features be $\mathbf{X}^{(l-1)} \in \mathbb{R}^{B \times C_{l-1} \times H_{l-1} \times W_{l-1}}$, where $B$ is the batch size, $C$ is the number of channels, and $H, W$ are the height and width of the feature map, respectively. The output features are $\mathbf{X}^{(l)} \in \mathbb{R}^{B \times C_{l} \times H_{l} \times W_{l}}$. The objective is to learn a lightweight mapping function such that:

\begin{equation} f_{\theta}(\mathbf{X}^{(l-1)}) \approx \mathbf{X}^{(l)} \end{equation}

To ensure that MeanFlow can accurately learn the intra-stage feature transformation, we employ a pre-trained ResNet model as a supervisory signal. The output features of each stage of the pre-trained model provide stable training targets for $z(0)$ and $z(1)$, enabling MeanFlow to learn a feature mapping that is equivalent to that of ResNet but more parameter-efficient. The features from each stage, after being mapped by MeanFlow, are passed sequentially, and the final classification results are output through a global average pooling layer and a fully connected layer. This design, while preserving the stage-wise structure of ResNet, significantly reduces the number of parameters and computational overhead within each stage through flow-field modeling.

To achieve the aforementioned objective, the MeanFlow mapping module consists of two key components, a dimension alignment module and a MeanFlow-based feature mapping module.

\subsubsection{Dimension Alignment}
ResNet employs a stage-wise design where the feature map size remains unchanged within each stage, which provides a natural structural basis for applying MeanFlow within stages. MeanFlow requires the source and target distributions to have the same dimension, and the input and output features within a ResNet stage satisfy this condition, allowing us to treat each stage as an independent flow mapping unit. However, there are significant dimension differences between different stages in ResNet. During stage transitions, the feature map size is halved while the number of channels is doubled, which hinders direct feature mapping learning across stages. To address this, we design a lightweight dimension alignment module as a transition layer between stages:
\begin{equation}
z_{\text{align}} = \text{ReLU} \left( \text{BN} \left( \mathbf{W}_{\text{align}} * \mathbf{X}^{(\ell-1)} \right) \right)
\end{equation}
where \(\mathbf{W}_{\text{align}}\) is a \(1 \times 1\) learnable convolutional kernel that can adaptively adjust the stride based on the spatial size relationship between the source and target features. The core role of this module is to align the source feature \(\mathbf{X}^{(\ell-1)}\) uniformly to the target dimension $C_\ell \times H_\ell \times W_\ell$ providing dimensionally consistent input for subsequent flow matching processes.

\subsubsection{Feature Mapping}
First, the dimension-aligned features are flattened into a two-dimensional matrix \(z \in \mathbb{R}^{(B \cdot H_\ell \cdot W_\ell) \times C_\ell}\). Subsequently, the evolution of the probability path over the time interval \(t \in [0,1]\) is defined and described by an ODE, as shown in Equation~\ref{eq:emc}.

Given a time parameter pair ($r, t$), where $r$ is the start time and $t$ is the end time ($0 \leq r < t \leq 1$) and a feature pair \((z_{\text{align}}, z_{\text{target}})\), the target velocity field is computed via the Jacobian-vector product (JVP) to accurately capture the nonlinear transformation in the feature space, as follows:
\begin{equation}
u_{\text{target}} = v - (t - r) \cdot \frac{\partial u_\theta}{\partial t}, \quad v = z_{\text{align}} - z_{\text{target}}
\end{equation}

Our training objective is the flow matching loss, defined by optimizing the prediction accuracy of the velocity field, as follows:\begin{equation}
\mathcal{L}_{\text{MF}} = \frac{1}{N} \sum_{i=1}^{N} \| u_\theta (z_i, t_i) - u_{\text{target}, i} \|_2^2 \end{equation} where \(u_\theta (z_i, t_i)\) is the velocity field predicted by the model, and \(u_{\text{target}, i}\) is the target velocity field.

We employ a differentiated training strategy. The first three stages utilize single-step training, whereas the fourth stage adopts a two-step training strategy. Specifically, the fourth stage decomposes the complete distribution transformation process into two sub-steps for learning. The first step learns the transformation from $t = 0$ to $t = 0.5$, and the second step learns the transformation from $t = 0.5$ to $t = 1$:
\begin{align}
z_{0.5}^{(4)} &= z_{0}^{(4)} + 0.5 \cdot u_{\theta}^{(4,1)}(z_{0}^{(4)}, 0, 0.5) \\
z_{1}^{(4)} &= z_{0.5}^{(4)} + 0.5 \cdot u_{\theta}^{(4,2)}(z_{0.5}^{(4)}, 0.5, 0.5)
\end{align}
During the inference phase, complex time parameter sampling is not required. By fixing the time parameters to $(r, t) = (0, 1)$, the feature mapping is directly accomplished via a single-step ODE solver:
\begin{equation}
z_{\text{mapped}} = z_{\text{align}} - u_{\theta}(z_{\text{align}}, 0, 1)
\end{equation}
The mapped flattened features are then reshaped into a tensor, which can replace the original features of stage $l$ for forward propagation, thereby achieving efficient end-to-end inference.

\subsubsection{Multi-stage Meta Model}
\label{sec:method313}
After completing the independent training of the MeanFlow modules for each stage, we construct the complete meta model architecture in two steps, namely cascading and fine-tuning.

Let the pre-trained ResNet contain \( L \) stages, and denote the MeanFlow mapping module of the \( l \)-th stage as \( M_l \). The cascaded meta model is defined as a multi-stage composite mapping:\begin{equation}
z_l = M_l(z_{l-1}), \quad \text{for} \quad l = 1, 2, \ldots, L
\end{equation} where each \( M_l \) incorporates both the dimension alignment module and the MeanFlow velocity field, with its parameters inherited from the independent training in Section~\ref{sec:method31}. This definition establishes an end-to-end mapping from the input image \( x \) to the prediction \( y \). However, since the modules were trained independently without considering the coupling effects between stages, direct cascading may lead to performance degradation due to feature distribution shift.

To mitigate the coupling error between modules, we perform end-to-end fine-tuning on the cascaded meta model. The optimization objective during fine-tuning is the classification cross-entropy loss. The fine-tuning strategy employs a small learning rate and freezes the parameters of the initial convolutional layers extracted from the pre-trained ResNet, optimizing only the parameters of the velocity field networks within each MeanFlow module. The resulting meta model after fine-tuning serves as the initial state for Deep Incubation.

\subsection{Selective Incubation Strategy}
\label{sec:method32}
Based on the meta model assembled in Section~\ref{sec:method313}, we adopt a selective incubation strategy to balance model compression and performance. This strategy is tailored to the parameter distribution of ResNet across stages. ResNet exhibits a pronounced parameter imbalance. As network depth increases, the number of feature channels grows stage by stage, causing parameters to concentrate in later stages. For example, in ResNet-50, although the numbers of residual blocks per stage are 3, 4, 6, and 3, respectively, the fourth stage often contains approximately 60\% of the network’s total parameters due to the larger channel width.

\begin{figure*}[t]
  \centering
  \includegraphics[width=0.95\linewidth]{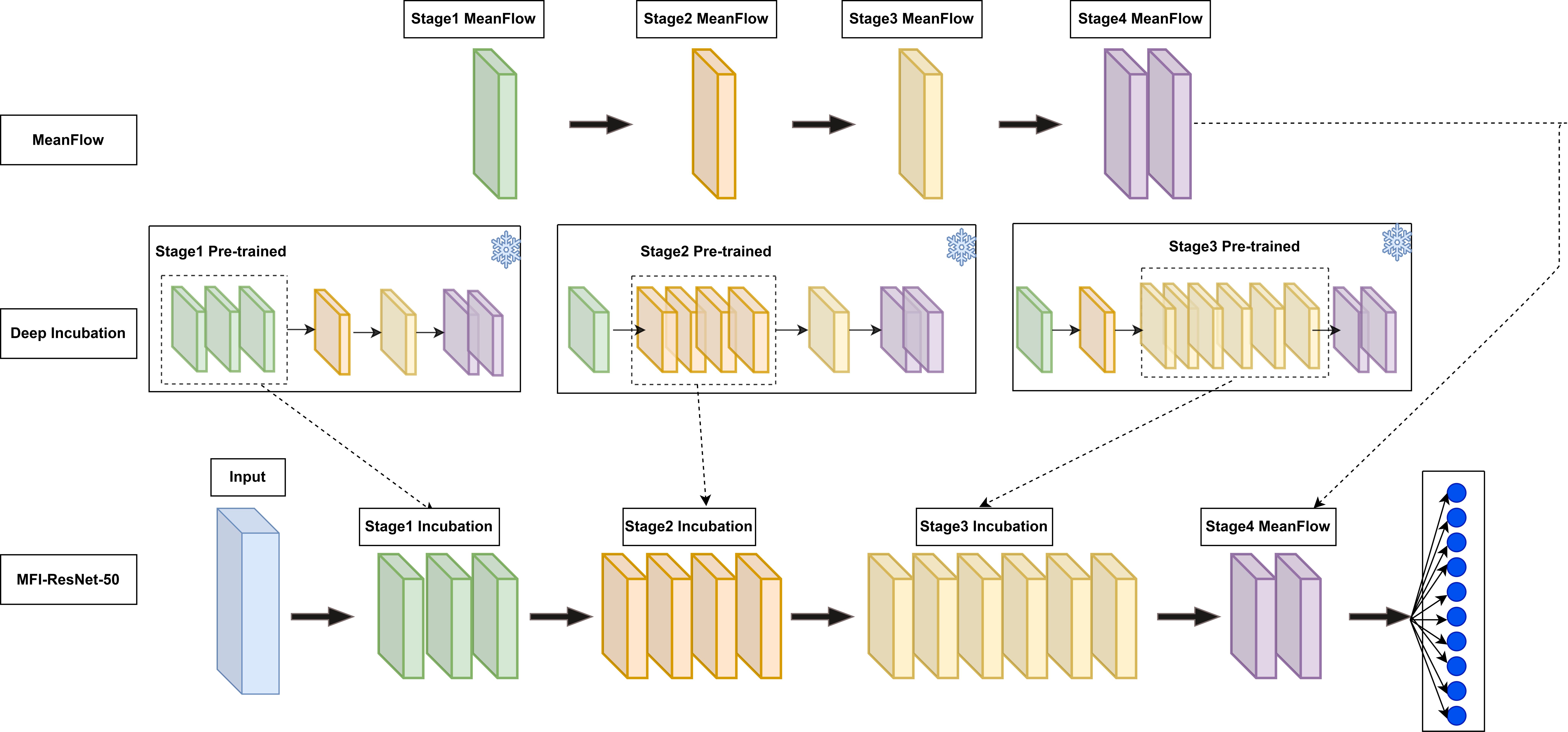}
   \caption{Selective incubation strategy for MFI-ResNet construction. \textbf{Top:} Four independent MeanFlow modules are trained to learn stage-wise feature mappings from pre-trained ResNet. \textbf{Middle:} Deep Incubation process where each stage (1-3) is independently replaced with corresponding pre-trained ResNet residual blocks while keeping other stages frozen. \textbf{Bottom:} Final MFI-ResNet-50 architecture combines incubated stages 1-3 with the retained two-layer MeanFlow module for stage 4, followed by global fine-tuning to achieve optimal performance.}
   \label{fig:incubation}
\end{figure*}

Leveraging this observation, we perform Deep Incubation on the first three stages to restore discriminative features, while retaining two MeanFlow modules in the fourth stage to preserve compression gains. Considering that a single-step MeanFlow may be insufficient for modeling high-dimensional, complex distributions, stage~4 employs a two-layer MeanFlow module that performs a two-step continuous transformation to more precisely characterize the evolution of feature distributions.

For a stage \(l\) scheduled for incubation, the procedure consists of two steps. The first step is module replacement, and the second step is end-to-end training. First, we replace the MeanFlow module \(\mathrm{MF}_l\) in the meta model with the pre-trained residual block sequence of stage \(l\) from ResNet, keeping the pre-trained weights as initialization. Next, we fine-tune the resulting hybrid architecture end-to-end, updating only the parameters of the currently incubated stage while freezing the others. The optimization objective is the standard cross-entropy loss for classification.

We independently execute the aforementioned replacement and training process for the first three stages. Each stage is incubated based on the meta model independently, and the trained weights for that stage are saved upon completion of incubation. After incubating the three stages, we obtain a hybrid architecture whose first three stages are the incubated counterparts of the corresponding ResNet stages, while the fourth stage remains a two-layer MeanFlow module. Finally, the entire model undergoes end-to-end fine-tuning, during which all parameters are updatable, and the optimization objective remains the cross-entropy loss. This global fine-tuning enables collaborative optimization across all stages, further enhancing the overall performance of the model.
\section{Experiment}
\subsection{Experimental Settings}

\begin{table*}[t]
\centering
\caption{Stage-wise parameter distribution in ResNet-34 and ResNet-50 architectures}
\label{tab:config}
\begin{tabular}{lcccccc}
\toprule
\multirow{2}{*}{Stage} & \multicolumn{3}{c}{ResNet-34} & \multicolumn{3}{c}{ResNet-50} \\
\cmidrule(lr){2-4} \cmidrule(lr){5-7}
 & \#Blocks & Channels & Params(M) & \#Blocks & Channels & Params(M) \\
\midrule
Stage1 & 3$\times$BasicBlock & 64 & 0.22 & 3$\times$Bottleneck & 256 & 0.22 \\
Stage2 & 4$\times$BasicBlock & 128 & 1.12 & 4$\times$Bottleneck & 512 & 1.22 \\
Stage3 & 6$\times$BasicBlock & 256 & 6.82 & 6$\times$Bottleneck & 1024 & 7.10 \\
Stage4 & 3$\times$BasicBlock & 512 & 13.11 & 3$\times$Bottleneck & 2048 & 14.96 \\
\bottomrule
\end{tabular}
\end{table*}
\textbf{Datasets} We conduct extensive experiments on two widely used image classification benchmarks, CIFAR-10 and CIFAR-100. CIFAR-10 consists of 60,000 $32\times32$ color images across 10 classes (50,000 for training and 10,000 for testing). CIFAR-100 contains the same number of images but with 100 fine-grained classes. Both datasets are normalized using channel-wise mean and standard deviation.

\noindent \textbf{Network Architectures} We evaluate our approach on two representative ResNet variants, ResNet-34 and ResNet-50. ResNet-34 employs BasicBlock with expansion factor 1, while ResNet-50 uses Bottleneck blocks with expansion factor 4. The detailed configurations of the four convolutional stages are shown in Table~\ref{tab:config}.

\noindent \textbf{Implementation Details} We use standard data augmentation including random cropping and random horizontal flipping. All images are normalized using channel-wise mean and standard deviation. In the MeanFlow module, we adopt the logit-normal sampling strategy, where the time parameters $r$ and $t$ are in $[0,1]$, generated via sigmoid function applied to a normal distribution with mean -0.4 and variance 1.0. During training, 75\% of the samples use $r=t$ representing the ODE endpoint, while 25\% of the samples use $r \neq t$. We employ AdamW optimizer with weight decay 0.01 and cosine annealing schedule. For the MeanFlow mapping stage, the learning rate is set to $2 \times 10^{-4}$ with 300 training epochs. For the selective incubation stage, each stage is trained independently for 200 epochs with learning rate 0.001. For the global fine-tuning stage, all parameters are optimized for 100 epochs with the same learning rate. For the meta model assembly and incubation fine-tuning stages, the learning rate is set to 0.001 with 100 training epochs. Label smoothing with $\varepsilon=0.1$ is applied to improve generalization. All models are trained with batch size 128 per GPU on 9 NVIDIA RTX 3090 GPUs using PyTorch DistributedDataParallel (DDP).

\subsection{Stage-wise MeanFlow Training}
We first validate the effectiveness of the stage-wise training strategy by examining the stand-alone performance of the MeanFlow modules and their integration within the cascaded meta model. Specifically, we assess: (1) whether each MeanFlow module can accurately learn an explicit mapping from shallow to deep features; and (2) whether cascading multiple modules yields strong classification performance.

According to the requirements outlined in Section~\ref{sec:method31}, we independently trained four MeanFlow modules for the four stages of ResNet, with each module learning to map the feature space from the preceding pre-trained layer to the corresponding subsequent pre-trained layer.

Next, we integrate the four trained MeanFlow modules into a cascaded architecture and perform end-to-end fine-tuning for the classification task. Table~\ref{tab:performance} compares the meta models with pre-trained ResNet baselines and ResNet-18. Experimental results show that the meta models based on ResNet-34 and ResNet-50 utilize approximately 5M parameters on both datasets, achieving a 75\%--78\% parameter reduction compared to their corresponding pre-trained baselines. The accuracy drop is only 2.73\%--3.21\%. Notably, the ResNet-34–based meta model achieves comparable or even better accuracy than ResNet-18 on both CIFAR-10 and CIFAR-100, while using less than half of its parameters. These results validate the feasibility of replacing ResNet's implicit transformations with explicit generative mappings, demonstrating that MeanFlow modules successfully capture the core evolutionary patterns of ResNet's intermediate features while maintaining strong discriminative performance with significantly fewer parameters.

\begin{table}[t]
\centering
\small % 缩小表格字体
\setlength{\tabcolsep}{2pt} % 缩小列间距
\renewcommand{\arraystretch}{1.1} % 增加行高
\caption{Performance comparison between meta models and pre-trained ResNet baselines on CIFAR-10 and CIFAR-100 datasets. Parameters in millions (M) and accuracy in percentage (\%). \textbf{Bold} indicates meta model results.}
\label{tab:performance}
\begin{tabular}{ccccc}
\toprule
\multirow{2}{*}{Model} & \multicolumn{2}{c}{CIFAR-10} & \multicolumn{2}{c}{CIFAR-100} \\
\cmidrule(lr){2-3} \cmidrule(lr){4-5}
 & Params (M) & Acc (\%) & Params (M) & Acc (\%) \\
\midrule
ResNet-18 & 11.17 & 92.14 & 11.21 & 71.55 \\[2pt]
Pre-trained (R-34) & 21.28 & 95.11 & 21.32 & 75.17 \\[2pt]
\textbf{Meta Model (R-34)} & \textbf{5.08} & \textbf{92.13} & \textbf{5.39} & \textbf{72.16} \\[2pt]
Pre-trained (R-50) & 23.51 & 95.34 & 23.69 & 75.80 \\[2pt]
\textbf{Meta Model (R-50)} & \textbf{5.11} & \textbf{92.84} & \textbf{5.43} & \textbf{72.59} \\
\bottomrule
\end{tabular}
\end{table}

\subsection{MFI-ResNet}

Having validated the effectiveness of the meta model, we further explore a Deep Incubation strategy that aims to transfer the explicit mapping capability of MeanFlow back to the ResNet residual structure, thereby improving inference efficiency while maintaining performance.

Table~\ref{tab:config} presents the parameter distribution across stages of ResNet-34. Stages~1-3 collectively account for only 38.4\% of the total parameters, while stage~4 contains 61.6\% of the parameters. This highly imbalanced parameter distribution indicates that the later stages carry the majority of the network's computational complexity. Accordingly, we choose to incubate stages~1-3 to recover discriminative features, while retaining the MeanFlow modules in stage~4 to preserve compression gains, thereby constructing a hybrid architecture.

We independently incubate stages~1-3, training each stage for 200 epochs. For each stage, we replace the MeanFlow module in the meta model with the corresponding ResNet residual block sequence and train the parameters of that stage end-to-end using the classification loss. After completing the three independent incubations, we cascade all incubated modules together with the stage~4 MeanFlow module to construct the hybrid MFI-ResNet, unfreeze all parameters, and perform 100 epochs of global fine-tuning.

\begin{table}[t]
\centering
\small
\setlength{\tabcolsep}{1.5pt}
\renewcommand{\arraystretch}{1.2}
\caption{Performance comparison of MFI-ResNet on CIFAR-10 and CIFAR-100 datasets. Parameters in millions (M) and accuracy in percentage (\%). Superscript numbers indicate the difference compared to corresponding pre-trained baselines.}
\label{tab:mfi-performance}
\begin{tabular}{lllll}
\toprule
\multicolumn{1}{c}{\multirow{2}{*}{Model}} & \multicolumn{2}{c}{CIFAR-10} & \multicolumn{2}{c}{CIFAR-100} \\
\cmidrule(lr){2-3} \cmidrule(lr){4-5}
 & \multicolumn{1}{c}{Params(M)} & \multicolumn{1}{c}{Acc(\%)} & \multicolumn{1}{c}{Params(M)} & \multicolumn{1}{c}{Acc(\%)} \\
\midrule
Pre-trained (R-34) & 21.28 & 95.11 & 21.32 & 75.17 \\
MFI-ResNet-34 & 12.38$^{\textcolor{blue}{\downarrow 8.90}}$ & 95.32$^{\textcolor{red}{\uparrow 0.21}}$ & 12.59$^{\textcolor{blue}{\downarrow 8.73}}$ & 75.21$^{\textcolor{red}{\uparrow 0.04}}$ \\
Pre-trained (R-50) & 23.51 & 95.34 & 23.69 & 75.80 \\
MFI-ResNet-50 & 12.62$^{\textcolor{blue}{\downarrow 10.89}}$ & 95.56$^{\textcolor{red}{\uparrow 0.22}}$ & 12.88$^{\textcolor{blue}{\downarrow 10.81}}$ & 75.93$^{\textcolor{red}{\uparrow 0.13}}$ \\
\bottomrule
\end{tabular}
\end{table}

Table~\ref{tab:mfi-performance} presents a performance comparison between MFI-ResNet and pre-trained ResNet baselines on the CIFAR-10 and CIFAR-100 datasets. The experimental results validate the effectiveness of our proposed compression-expansion strategy. On CIFAR-10, MFI-ResNet-34 achieves a 41.8\% parameter reduction compared to the pre-trained ResNet-34 baseline while simultaneously improving accuracy by 0.21\%. MFI-ResNet-50 achieves a 46.3\% parameter reduction compared to the pre-trained ResNet-50 baseline with an accuracy improvement of 0.22\%. On CIFAR-100, MFI-ResNet-34 and MFI-ResNet-50 achieve parameter reductions of 40.9\% and 45.6\%, respectively, with accuracy improvements of 0.04\% and 0.13\%.

These results demonstrate that, with the selective incubation strategy, MFI-ResNet achieves better classification performance while retaining the substantially lower parameter count of the meta model compared to the baseline ResNet, by recovering critical discriminative features. Notably, all MFI-ResNet variants achieve performance improvements while substantially reducing parameter counts, validating the synergistic effect between the compression phase and the expansion phase. These findings indicate that introducing generative flow-field models can achieve more efficient feature transformations than conventional ResNet, providing valuable empirical support for exploring the potential connections between discriminative learning and generative modeling at the feature transformation level.

\section{Conclusion}
In this work, we presented MeanFlow-Incubated ResNet (MFI-ResNet), which optimizes ResNet architectures by applying generative flow-field modeling under a unified compression-expansion strategy. Starting from the ODE view of ResNet, where multi-step residual connections correspond to discrete integration of an underlying velocity field, we compress each stage by replacing its multi-layer residual accumulation with one or two MeanFlow modules that model the mean velocity field, yielding a lightweight meta model with substantially fewer parameters. In the expansion phase, a selective incubation strategy exploits the stage-wise parameter imbalance of ResNet by restoring shallow stages with their incubated ResNet counterparts while retaining a MeanFlow-based final stage to preserve compression gains. Experiments on CIFAR-10 and CIFAR-100 show that MFI-ResNet reduces parameters by roughly 46\% compared with ResNet-50 while slightly improving classification accuracy. These results suggest that generative flow-field modeling, grounded in the ODE interpretation of residual networks, provides a promising perspective for understanding residual feature transformations in modern deep architectures.

{
    \small
    \bibliographystyle{ieeenat_fullname}
    \bibliography{main}
}


\end{document}